\documentclass[letterpaper, 10 pt, conference]{ieeeconf}  

\IEEEoverridecommandlockouts                              

\usepackage{graphicx}
\usepackage[ruled, lined, linesnumbered, commentsnumbered, longend]{algorithm2e}

\usepackage{etex}

\usepackage{amssymb}
\usepackage{mathrsfs}
\usepackage{amsmath}

\usepackage{amsthm}

\usepackage{multirow}

\usepackage[sans]{dsfont}
\usepackage{stmaryrd}
\usepackage{pifont}
\usepackage{wasysym}

\usepackage{array}
\usepackage{url}
\usepackage{colortbl}
\usepackage{caption}
\usepackage{subcaption}
\usepackage{color}
\usepackage{diagbox,pict2e}

\usepackage{mathtools}

\newtheorem{theorem}{Theorem}

\newtheorem{lemma}{Lemma}
\newtheorem{corollary}[theorem]{Corollary}

\newtheorem*{remark*}{Remark}

\def\cA{\mathcal A}

\def\cG{\mathcal G}

\def\cN{\mathcal N}

\def\cS{\mathcal S}

\def\RR{\mathbb R}
\def\RR{\mathbb R}

\def\bx{\mathbf{x}}

\def\bz{\mathbf{z}}

\def\bzero{\mathbf{0}}

\newcommand{\raf}[1]{(\ref{#1})}

\title{\LARGE \bf
 Dual Formulation for Chance Constrained Stochastic Shortest Path with Application to Autonomous Vehicle Behavior Planning} 
 \author{Rashid Alyassi, Majid Khonji 
 \thanks{
 Rashid Alyassi and Majid Khonji are with EECS Department,  Khalifa University, Abu Dhabi, UAE (Emails: \texttt{\{rashid.alyassi, majid.khonji\}@ku.ac.ae}).
 This work was supported by the Khalifa University of Science and Technology under award references CIRA-2019-049, KKJRC-2019-Trans1 and KUCARS.}
 }

\begin{document}

\maketitle
\thispagestyle{empty}
\pagestyle{empty}

\begin{abstract}
Autonomous vehicles face the problem of optimizing the expected performance of subsequent maneuvers while bounding the risk of collision with surrounding dynamic obstacles. These obstacles, such as agent vehicles, often exhibit stochastic transitions that should be accounted for in a timely and safe manner.
The Constrained Stochastic Shortest Path problem (C-SSP) is a formalism for planning in stochastic environments under certain types of operating constraints.  While C-SSP allows specifying constraints in the planning problem, it does not allow for bounding the probability of constraint violation, which is desired in safety-critical applications. 
This work's first contribution is an exact integer linear programming formulation for Chance-constrained SSP (CC-SSP) that attains deterministic policies. Second, a randomized rounding procedure is presented for stochastic policies. Third, we show that the CC-SSP formalism can be generalized to account for constraints that span through multiple time steps.
Evaluation results show the usefulness of our approach in benchmark problems compared to existing approaches.
\end{abstract}

\section{Introduction}

The Markov Decision Process (MDP) \cite{howard1960dynamic} is a widely used model for planning under uncertainty. An MDP consists of states, actions, a stochastic transition function, a utility function, and an initial state. A solution of MDP is a policy that maps each state to an action that maximizes the global expected utility.  Stochastic Shortest Path (SSP) \cite{bertsekas1991analysis}  is an MDP with non-negative utilities and a minimizing objective. The problem has interesting structures and can be formulated with a dual linear programming (LP) formulation \cite{d1963probabilistic} that can be interpreted as a minimum cost flow problem. Moreover,  SSP  has many heuristics-based algorithms \cite{bonet2003labeled, hansen2001lao} that utilize admissible heuristics to guide the search without exploring the whole state space.

Besides, Constrained SSP (C-SSP)\cite{altman1999constrained} provides the means to add mission-critical requirements while optimizing the objective function. Each requirement is formulated as a  {\em budget constraint} imposed by a non-replenishable resource for which a bounded quantity is available during the entire plan execution. Resource consumption at each time step reduces the resource availability during subsequent time steps (see ~\cite{de2021constrained} for a detailed discussion).
A stochastic policy of C-SSP is attainable using several efficient algorithms (e.g., \cite{feinberg1996constrained}).  A heuristics-based search approach in the dual LP can further improve the running time for large state spaces \cite{trevizan2016heuristic}. For deterministic policies, however, it is known that C-SSP is NP-Hard~\cite{feinberg2000constrained}.

A special type of constraint occurs when we want to bound the probability of constraint violations by some threshold $\Delta$, which is often called {\em chance constraint}. To simplify the problem,  \cite{dolgov2003approximating} proposes approximating the constraint using Markov's inequality, which converts the problem to C-MDP. Another approach \cite{de2017bounding} applies Hoeffding's inequality on the sum of independent random variables to improve the bound. Both methods provide conservative policies that respect safety thresholds but at the expense of the objective value.

In the partially observable setting, the problem is called  Chance Constrained Partially Observable MDP (CC-POMDP). Several algorithms address CC-POMDP under {\em local} risk constraints, where risk is dependent only on the state, and the chance constraint is defined as the probability of failure in {\em any} state during execution~\cite{santana2016rao,khonji2019approximability}. However, due to partial observability, these methods require an enumeration of histories, making the solution space exponentially large with respect to the planning horizon. To speed up the computation, \cite{sungkweon2021ccssp} provides an {\em anytime} algorithm using a Lagrangian relaxation method for CC-SSP and CC-POMDP that returns feasible sub-optimal solutions and gradually improves the solution's optimality when sufficient time is permitted. Unfortunately, the solution space is represented as an And-Or tree, similar to other CC-POMDP methods, causing the algorithm to be slow in the planning horizon.

 \begin{figure}[t!]
    \centering 
    \includegraphics[trim={0 11mm 0 0},clip,width=\linewidth] {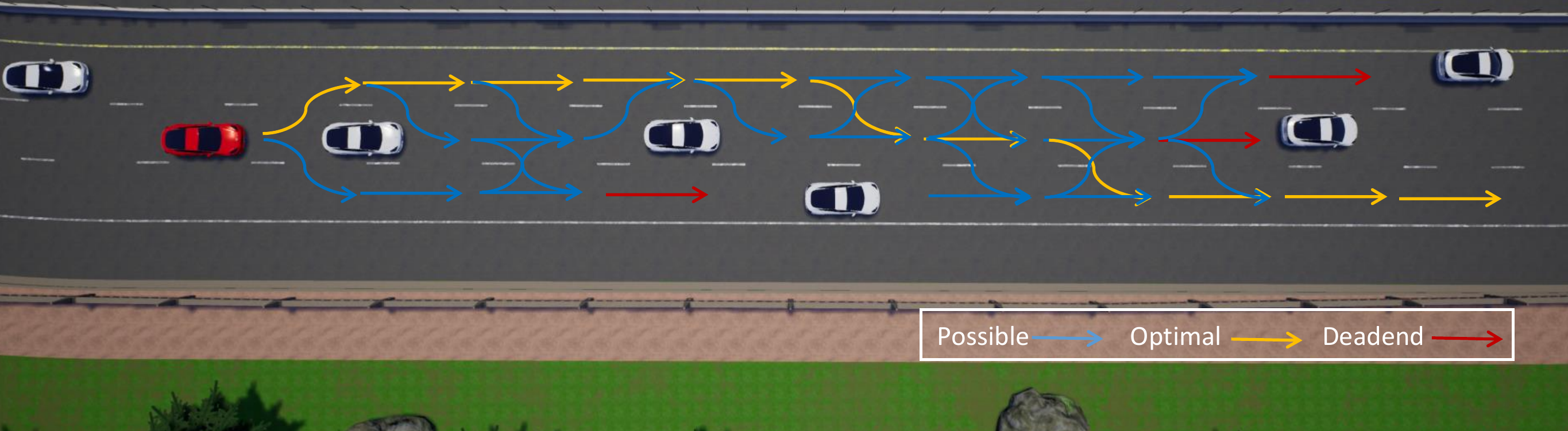}
    \caption{The red vehicle is the ego-vehicle (AV), while the white vehicles are human-driven vehicles (HVs). Arrows show potential maneuvers for the ego-vehicle.  }
    \label{fig:hw}\vspace{-15pt}
\end{figure}
Behavior planning for Autonomous Vehicles (AVs) has been extensively studied in deterministic environments (see e.g., \cite{wei2014behavioral}). One of the primary sources of uncertainty arises from drivers' intentions, i.e.,  potential maneuvers of agent vehicles in the scene \cite{huang2019online}. An effective planner should optimize the maneuvers (say, minimize total commute time) while bounding collision probability below some threshold. More recently, \cite{huang2018hybrid,huang2019online} show that such a plan is achievable under more detailed settings. Essentially, they modeled the problem as a CC-POMDP, where actions resemble (deterministic) ego-vehicle maneuvers, and observations resemble predictions of agent vehicle maneuvers (see Fig.~\ref{fig:hw}). In fact, the problem can be equivalently modeled as CC-SSP, with transitions used for joint actions and agent vehicle predictions. Unlike CC-POMDP, our CC-SSP formulation scales {\em polynomially} in the planning horizon resulting in a significant performance gain.

This paper's contribution is threefold. First, an {\em exact} integer linear programming (ILP) formulation is provided for CC-SSP that obtains deterministic policies in the dual space. The chance-constrained formulation is based on the notion of {\em execution risk}  introduced in the literature in the context of CC-POMDP  \cite{santana2016rao} under local risk constraints. 
Second, we generalize upon local CC-SSP constraints to {\em global} constraints.
Unlike \cite{sungkweon2021ccssp}, our formulation does not expand all possible histories, which entails exponential growth in solution space. Our formulation is also a departure from existing approximate approaches \cite{dolgov2003approximating,de2017bounding} that have no bounded performance with respect to the objective value.
Third, we present a randomized rounding algorithm for the CC-SSP that provides a close-to-optimal solution in practice within a reasonable time.
Both stochastic and deterministic approaches are evaluated under two benchmark problems, including a highway behavior planning problem for autonomous vehicles (Fig.~\ref{fig:hw}).

\section{CC-SSP with Local Risk Constraints}
In this section, we present (A) the problem definition of the C-SSP and CC-SSP, (B) a method for computing the {\em execution risk} for the CC-SSP, (C) the ILP formulation for the CC-SSP, and (D) a Randomized Rounding algorithm for the CC-SSP. 
\subsection{Problem Definition}
We provide formal definitions for C-SSP and CC-SSP as follows.
	A fixed-horizon constrained stochastic shortest path (C-SSP) is a tuple  \mbox{$ \langle \cS, \cA, T, U, s_0, h,(C^j, P^j)_{j\in \cN} \rangle$}, 
	\begin{itemize}
    \item 
$\cS$ and $\cA$ are finite sets of discrete {states} and {actions}, respectively;
	\item
$T: \cS \times\cA\times \cS \rightarrow [0,1]$ is a probabilistic {transition function} between  states, 
$T(s,a,s') = \Pr(s' \mid a,s)$, where $s,s' \in \cS$ and $a\in \cA$;
	\item  $U: \cS \times \cA\rightarrow \RR_+$ is a non-negative {utility function};	
	\item $s_0$ is an initial state;
	\item $h$ is the planning horizon; 
	\item $C^j: \cS \times \cA\rightarrow \RR_+$ is a non-negative {cost function}, with respect to criteria $j\in \cN$, where $\cN$ is the index set of all risk criteria; 	
	\item  $P^j \in \RR_+$ is a positive upper bound on the cost criteria $j \in  \cN$.
	\end{itemize}
A {\em deterministic} policy $\pi(\cdot,\cdot)$ is a function that maps a state and time step into an action, $\pi:\cS\times \{0,1,...,h-1\} \rightarrow \cA$. A {\em stochastic} policy $\pi: \cS \times \{0,1,...,h-1\}\times \cA \rightarrow [0,1]$ is defined as a distribution over actions from a given state and time. For simplicity, we write $\pi(S_t)$ to denote $\pi(S_t,t)$, and $\pi(s_k,a)$ to denote stochastic policy $\pi(s_k,k,a)$. A {\em run} is a sequence of random states  $S_0, S_1,\ldots, S_{h-1}, S_h$ that result from executing a policy, where $S_0 = s_0$ is known. 

The objective is to compute a policy that maximizes (resp. minimizes) the expected utility (resp. cost)  while satisfying all the constraints.
More formally,
\begin{align}
\textsc{(C-SSP)}\quad&\quad\max_{\pi} \mathbb{E}\Big[\sum_{\mathclap{t=0}}^{h-1} U(S_{t}, \pi(S_{t}) )  \Big] \label{obj1}\\
\text{Subject to }\quad &  \mathbb{E} \Big[\sum_{t=0}^{h-1} C^j(S_t, \pi(S_t))\mid \pi \Big] \le \ P^j, \quad j \in \cN.\notag
\end{align}

	A fixed-horizon chance-constrained stochastic shortest path (CC-SSP) problem is formally defined  as a tuple 
$M=\langle \cS, \cA, T,U, s_0, h, (r^j, \Delta^j)_{j\in \cN} \rangle,$ where $\cS, \cA, T, U, s_0, h, \cN$ are defined as in C-SSP, 
	\begin{itemize}
\item $r^j: \cS \rightarrow [0,1]$ is the probability of failure  at a given state according to risk criterion $j$;
\item $\Delta^j$ is the corresponding risk budget, a threshold on the probability of failure over the planning horizon, for $j\in \cN$.
	\end{itemize}
Let $R^j(s)$ be a Bernoulli random variable for failure at state $s$ with respect to criterion $j\in \cN$, such that $R^j(s)=1$ if and only if it is in a risky state, and zero otherwise. For simplicity, we write $R^j(s)$ to denote $R^j(s)=1$.
The objective of {CC-SSP} is to compute a policy (or a conditional plan) $\pi$ that maximizes the cumulative expected utility while bounding the probability of failure at {\em any} time step throughout the planning horizon.
\begin{align}
\textsc{(CC-SSP)}\quad&\quad\max_{\pi} \mathbb{E}\Big[\sum_{\mathclap{t=0}}^{h-1} U(S_{t}, \pi(S_{t}) )  \Big] \\
\text{Subject to }\quad &  \Pr\Big(\bigvee_{t=0}^{h} R^j(S_t) \mid \pi \Big) \le \Delta^j, & j \in \cN. \label{con1}
\end{align}
We refer to such constraints as {\em local} since failure could occur entirely in a single step (i.e., $R^j(S_t) = 1$). Sec.~\ref{sec:global} considers failure as accumulative throughout an entire run. In both cases, the constraint bound the probability of failure.

\begin{figure}[ht]
     \centering 
     \includegraphics[width=200pt]{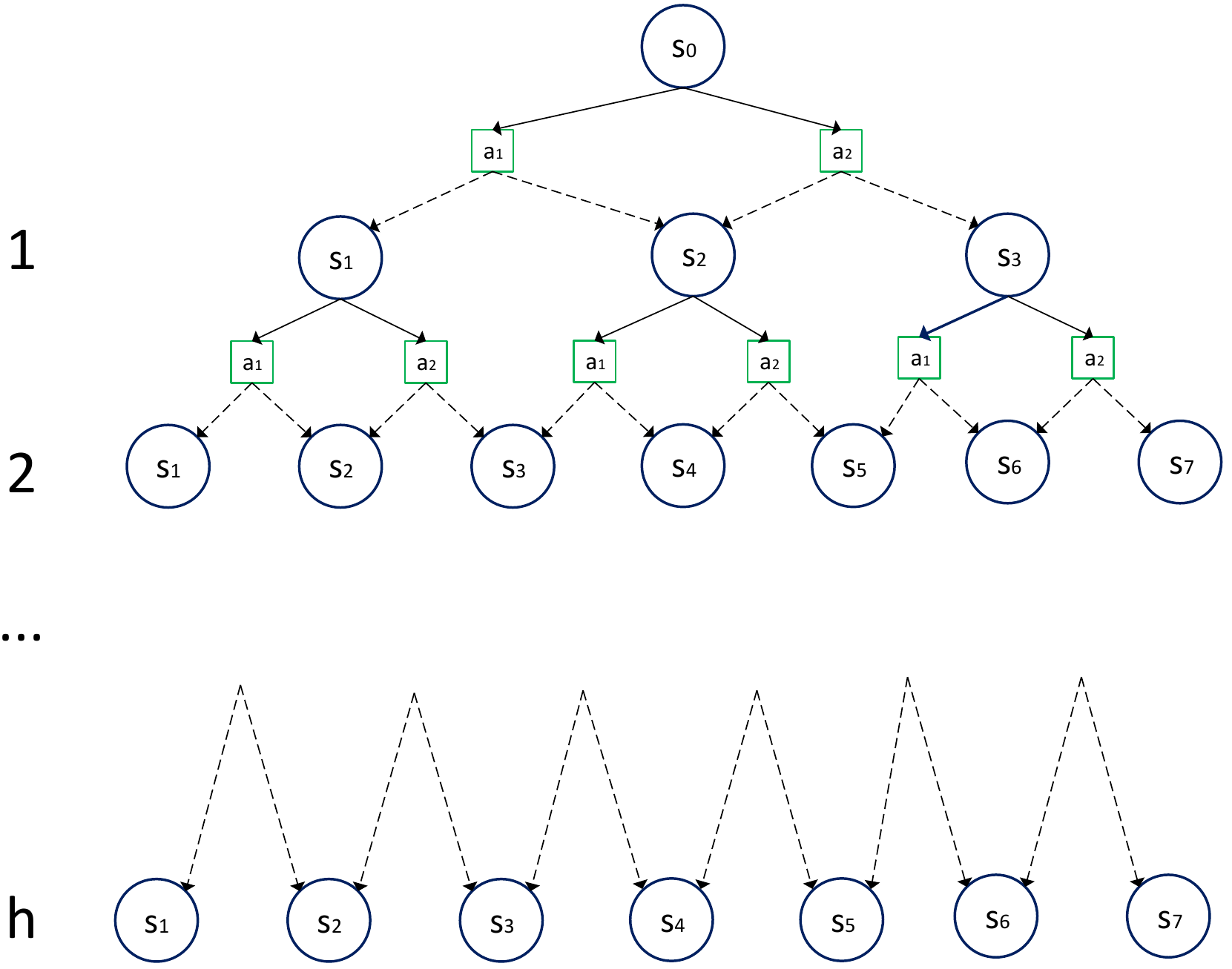}
     \caption{SSP graph}\vspace{-10pt}
     \label{fig:ccssp_graph}
 \end{figure}

The SSP problem and its constrained variants can be visualized by a Direct Acyclic Graph (DAG), where the vertices represent the states and their actions, and the edges are in two types. The edge between a state and an action represents the action's cost, while the edge between an action and a state represents the transition probability.
Thus, at depth $k$ all the states are reachable from previous actions taken at depth $k-1$, and each depth has a maximum of $|\cS|$ states (i.e, all the states). Fig.~\ref{fig:ccssp_graph} provides a pictorial illustration of the CC-SSP And-Or search graph. Not that unlike And-Or search trees obtained by history enumeration algorithms (see, e.g., \cite{sungkweon2021ccssp}), with such representation, a state may have multiple parents, leading to significant reduction in the search space.

\subsection{Execution Risk} 

Define the {\em execution risk} of a run at state $s_k$  as 
$$\textsc{Er}^j(s_k) :=  \Pr\Big(\bigvee_{t=k}^{h} R^j(S_t) \mid S_k = s_k\Big).$$ According to the definition, Cons.~\raf{con1} is equivalent to $\textsc{Er}^j(s_0) \le \Delta^j$. The lemma below shows that such constraint can be computed recursively. 

\begin{lemma}\label{lem:rec}
The execution risk can be written as 
\begin{align}
 &\textsc{Er}^j(s_k) = \sum_{s_{k+1}\in \cS}\sum_{a \in \cA} \Big( \textsc{Er}^j(s_{k+1}) \pi(s_k, a) \widetilde T^j(s_k, a, s_{k+1}) \Big)\notag \\
 &\qquad\qquad~~+  r^j(s_k),\qquad j \in \cN \label{eq:exec-risk}
\end{align}
where $\widetilde T^j(s_k, \pi(s_k), s_{k+1}):=  {T(s_k, \pi(s_k), s_{k+1})}(1-r^j(s_k))$. 
\end{lemma}
A proof for Lemma~\ref{lem:rec} is deferred to the appendix.

\subsection{ILP Formulation}

Define a variable $x_{s,k,a}^j  \in [0,1]$ for each state $s_k$ at time $k$, and action $a$, and $j \in \cN\cup \{0\}$ such that
\begin{align}
&\sum_{a \in \cA}  x_{s,k,a}^j = \sum_{s_{k-1} \in \cS} \sum_{a \in \cA} x_{s,k-1,a}^j \widetilde T^j(s_{k-1}, a, s_{k}), \notag \\& \hspace{40mm} k = 1,...,h-1,  s_k \in \cS, \label{conf1}\\
&\sum_{a \in \cA} x^j_{s,0,a}  = 1. \label{conf2}
\end{align}
In fact, the above {\em flow} constraints for $j=0$ are the standard dual-space constraints for SSP~\cite{10.5555/2074158.2074203}. 
According to the above constraints, we can rewrite the execution risk at $s_0$, defined by Eqn.~\raf{eq:exec-risk}, according to the following key result.
\begin{theorem}\label{lem1}
Given a conditional plan $\bx$ that satisfies Eqn.~\raf{conf1}-\raf{conf2}, the execution risk can written as linear function of $\bx$,
\begin{align}\label{er_eq}
&\textsc{Er}^j(s_0) =  \sum_{k = 1}^{h}\sum_{s_{k-1}\in \cS}\sum_{a \in \cA} \sum_{s_k\in \cS}  r^j(s_{k}) x^j_{s,k-1,a} \widetilde T^j(s_{k-1}, a, s_{k}) \notag \\
&\qquad \qquad  + r^j(s_0)
\end{align}
\end{theorem}
\vspace{-6pt}
\noindent
A proof for theorem~\ref{lem1} is provided in the appendix.

\vspace{6pt}

Let $\widetilde T^0(\cdot,\cdot, \cdot) := T(\cdot,\cdot, \cdot)$. The chance-constrained stochastic shortest path problem (CC-SSP) can be formulated by the following ILP:
\begin{align}
&\textsc{(CC-SSP-ILP)}\quad\max_{\bx,\bz} \sum_{k=0}^{h-1}\sum_{s_k \in \cS, a \in \cA} x^0_{s,k,a} U(s_k, a) \label{cc-ssp-ilp}\\
&\text{Subject to }\quad\notag\\
&\sum_{a \in \cA}  x_{s,k,a}^j = \sum_{s_{k-1} \in \cS} \sum_{a \in \cA} x_{s,k-1,a}^j \widetilde T^j(s_{k-1}, a, s_{k}), \notag\\&\hspace{20mm} k = 1,...,h-1,  s_k \in \cS, j \in \cN \cup \{0\},\\
&\sum_{a \in \cA} x^j_{s,0,a}  = 1, \qquad j\in \cN \cup \{0\}, \\
&\sum_{k = 0}^{h-1}\sum_{s_{k}\in \cS}\sum_{a \in \cA} \sum_{s_{k+1}\in \cS}  r^j(s_{k+1}) x^j_{s,k,a} \widetilde T^j(s_k, a, s_{k+1}) \notag\\&\hspace{35mm} \le \Delta^j - r^j(s_0), \qquad j\in \cN, \label{riskcons} \\
&\sum_{a\in \cA} z_{s,k,a}\le 1, \qquad k=0,...,h-1, s_k \in \cS,\label{conzone}\\
& x^j_{s,k,a} \le z_{s,k,a}, \quad \forall j \in \cN\cup\{0\}, k=0,...,h-1, s_k \in \cS,\label{conzbind}\\
&z_{s,k,a}\in \{0,1\}, \quad x^j_{s,k,a} \in [0,1], \notag\\&\hspace{25mm} \forall j \in \cN\cup\{0\}, k=0,...,h-1, s_k \in \cS.
\end{align}
Cons~\raf{riskcons} follows directly from Theorem~\ref{lem1}.
The variable $z_{s,k,a}$ is used to bind the actions of all flows. Thus, for each constraint criterion $j$, Cons.~\raf{conzbind} ensures that for a given state, the same action is selected across all flows. Since, $z_{s,k,a}\in \{0,1\}$, Cons.~\raf{conzone}
ensures at most one deterministic action at each node.


\subsection{CC-SSP Randomized Rounding}
The CC-SSP is an NP-hard problem \cite{khonji2019approximability}, hence an exact solver such as the ILP discussed above can have, in the worst case, an exponential running time. Thus, we utilize a randomized rounding algorithm as a heuristic to quickly obtain (potentially) sub-optimal solutions for the CC-SSP. As we show in our experiments, the algorithm attains close-to-optimal solutions in practice in reasonable running time.
The algorithm is a probabilistic algorithm that utilizes a relaxed LP solution of the problem and rounds it to get an integral solution. Note that a naive rounding procedure only satisfies the constraints at expectation, which is not the case with chance constraints.
An LP formation is based on the ILP formation defined earlier (CC-SSP-ILP), with the $z$ variable redefined as a continuous variable $z_{s,k,a}\in [0,1]$. The randomized rounding algorithm rounds each $z$ value per state with a probability proportional to its value. The rounding process is iterated until a feasible solution that satisfies all the constraints is acquired. The pseudocode is provided in Algorithm~\ref{rounding}.

\begin{algorithm}[ht]
\footnotesize
\caption{\texttt{Rounding}$[\bx,\bz]$}
\SetKwInOut{input}{Input}
\SetKwInOut{output}{Output}
\input{An LP feasible solution $\bx,\bz$}
\output{A policy $\hat \bz$ }
$\hat{\bz} \leftarrow \bzero$;\\
$\hat{\bx} \leftarrow \bzero$;\\
\While{$\exists j$ such that $\hat{x}^j$ is {\bf not} feasible with respect to Cons.~\raf{riskcons}}{
\For{$s_k \in \cS$, $k=0,...,h-1$}{
\If{$\sum_{a\in \cA}z_{s,k,a}\neq 0$}{
Choose $\hat{a}\in \cA$ with probability $\frac{z_{s,i,\hat{a}}}{\sum_{a\in \cA}z_{s,k,a}}$ \\
$\hat{z}_{s,k,\hat{a}} \leftarrow \sum_{a\in \cA}z_{s,k,a}$ \\
$\hat{z}_{s,k,a} \leftarrow 0, \qquad \forall a\in \cA$, $a\neq \hat a$\\
$\hat{x}^j_{s,k,\hat{a}} \leftarrow \sum_{a\in \cA}x_{s,k,a},\quad \forall j \in \cN$  \\
$\hat{x}^j_{s,k,a} \leftarrow 0,$ for $a\neq \hat a, \quad \forall a\in \cA, \quad \forall j \in \cN$  \\
Scale $x^j_{s,k+1,a}$ and  $z_{s,k+1,a}$, $s_{k+1}\in \cS, a \in \cA$ based on $\hat x^j_{s,k,\hat a}$ and $\hat{z}_{s,k,\hat a}$, respectively 
}
} 
%
}
\Return $\hat{z}$  
\label{rounding}
\end{algorithm}

\section{CC-SSP with Global Risk Constraints}\label{sec:global}
We consider {\em global} CC-SSP (GCC-SSP) defined as a tuple $M=\langle \cS, \cA, T,U, C, s_0, h, \cN, P^j, \Delta^j \rangle,$ where
	\begin{itemize}
	\item 
$\cS, \cA, T,U, s_0, h, \cN, \Delta^j $ are defined as in the CC-SSP;
	\item
$C^j: \cS \times \cA\rightarrow \RR$ is a secondary cost function,  for $j\in \cN=\{1,...,n\}$;	
\item $P^j$ is an upper bound on the cumulative cost, and $\Delta^j$ is the corresponding risk budget, a threshold on the probability that the cumulative cost function exceeds the upper bound over the planning horizon.
	\end{itemize}
    The objective of {GCC-SSP} is to compute a policy $\pi$ that satisfies,
    \begin{align}
\textsc{(GCC-SSP)}\quad&\quad\max_{\pi} \mathbb{E}\Big[\sum_{\mathclap{t=0}}^{h-1} U(S_{t}, \pi(S_{t}) )  \Big] \label{obj}\\
\text{Subject to }\quad &  \Pr\Big( \sum_{t=0}^{h-1} C^j(S_t, \pi(S_t)) > P^j\mid \pi \Big) \le \Delta^j, \notag\\
&\hspace{44mm} j \in \cN. \label{gcon1}
\end{align}
Note that the constraint in CC-SSP is local to each round, whereas in GCC-SSP, the constraint is dependant on the whole run, hence the name global. 
\subsection{Reduction to CC-SSP}
One way to solve a GCC-SSP instance $M$  is by reducing to a CC-SSP instance $M'$, defined by augmenting the state space to include all possible values of $g^j(k) := \sum_{t=0}^{k}  C^j(s_t, \pi(s_t))$, for $k=0,...,h-1$, and  $s_t\in\cS$. In other words, $\cS':=\cS\times_j \cG^j$, where $\cG^j$ is the set of all possible values of $g^j(k)$. Let $g:= (g^j)_{j\in\cN}$, $g^j \in \cG^j$. Unfortunately, the size of $\cG^j$ is exponentially large, hence, impractical for many application domains. We will show in subsection~\ref{sec:aug} how to reduce the state space to a polynomial size. Given the augmented state space $\cS'$, 
the risk probability of  state $\langle s, g \rangle\in \cS'$  as
\begin{equation}
r^j(\langle s, g \rangle) \leftarrow \left\{\begin{array}{l r}
1 & \text{if }  g^j >  P^j,\\
0 & \text{otherwise.}
\end{array}\right.
\end{equation}
The transition function  in the augmented space is defined as $T'(\langle s,g\rangle, a, \langle s',g' \rangle):=$
\begin{equation}
 \left\{ \begin{array}{l r}
T(s,a,s') & \text{if } \forall j, g'^j = g^j + C^j(s,a), \\
0 & \text{otherwise.}
\end{array}
\right.
\end{equation}
Clearly, the transition function is a valid probability distribution. Write $G^j(k):= \sum_{t=0}^{k}  C^j(S_t, \pi(S_t))$ to denote the random variable for the cumulative cost up until state $S_k$. The random variable takes values in $\cG^j$. Using the aforementioned reduction, we have
\begin{align*}
&\sum_{k=0}^{h-1} C^j(S_k, \pi(S_k)) > P^j \iff \bigvee_{k=0}^{h-1} (G^j(k) > P^j) \\& \iff \bigvee_{k=0}^{h-1} \bigvee_{g^j \in \cG^j: g^j > P^j} (\cG^j(k)=g^j)\\
&\iff \bigvee_{t=0}^{h}\bigvee_{s_t'\in \cS'} R^j(S_t'=s'_t)  \iff \bigvee_{t=0}^{h} R^j(S_t').
\end{align*}
Thus, the probability of these events are equivalent. Therefore,
solving CC-SSP instance $M'$ is equivalent to that of GCC-SSP $M$. Hence, we can use CC-SSP-ILP to solve GCC-SSP following the aforementioned reduction.
\begin{lemma}\label{lem:red}
GCC-SSP is reducible to CC-SSP.
\end{lemma}

\subsection{GCC-SSP via Resource Augmentation \label{sec:aug}}
In this subsection, we show how to discretize the sets $\cG^j$, denoted by $\hat \cG^j$, such that  $|\hat \cG^j|$ is polynomial in $h$ and $\frac{1}{\epsilon}$, where $\epsilon \in (0,1)$ is an input parameter, while slightly affecting the solution feasibility, i.e.,
\begin{equation}
\Pr\Big( \sum_{t=0}^{h-1}  C^j(S_t, \pi(S_t)) > (1+\epsilon) P^j \mid \pi \Big) \le \Delta^j. \label{eq:aug}
\end{equation}
Such model is often referred to as {\em resource augmentation}. In many application domains, a slight violation of resource capacity can be acceptable. If that's not the case, one can increase $P^j$ to account for $\epsilon$\footnote{Arguably,  increasing the bound $P^j$ changes the problem definition, where there may exist a better optimal for the original problem that is not attainable in the discretized version.}.  

The idea is  to discretized the cost function $C^j(\cdot, \cdot)$,  denoted by $\hat C^j(\cdot,\cdot)$, as follows. Let $C^j_{\max}:= \max_{s\in \cS, a\in \cA} C^j(s,a)$, and  $ K^j := \frac{\epsilon C^j_{\max}}{h}$. The discretized values are given by $\hat C^j(s,a):= \lceil \tfrac{C^j(s,a)}{K^j} \rceil$. 

Note that $\hat g^j(h-1):= \sum_{t=0}^{h-1} \hat C^j(S_t, \pi(S_t))  \le  \sum_{t=0}^{h-1} \lceil \tfrac{C^j_{\max}}{K^j} \rceil  = h \lceil \tfrac{h}{\epsilon} \rceil$. Hence, the descritized version of $\cG^j$, denoted by $\hat \cG^j$, takes (at most) the values $\hat \cG^j \subseteq \{0,1,....,  h \lceil \tfrac{h}{\epsilon}\rceil  \} $, which is polynomial in $\frac{1}{\epsilon}$ and $h$. Denote the {\em discretized} GCC-SSP by replacing Cons.~\raf{gcon1} of GCC-SSP by
\begin{equation}
\Pr\Big( K^j \sum_{t=0}^{h-1}  \hat  C^j(S_t, \pi(S_t)) > (1+\epsilon) P^j \mid \pi \Big) \le \Delta^j.  \label{eq:disc}
\end{equation} 
\begin{theorem}\label{thm:dis}
An optimal solution of the discretized GCC-SSP is  (super) optimal to GCC-SSP, but may violate Cons.~\raf{gcon1}  by  a factor of at most $1+\epsilon$, as shown by  Cons.~\raf{eq:aug}.
\end{theorem}

\noindent
\textbf{proof.}
We show that any feasible policy $\hat \pi$ for the {\em discretized} GCC-SSP  satisfies Eqn.~\raf{eq:aug}, and that any feasible policy $\pi'$ of  GCC-SSP satisfies Eqn.~\raf{eq:disc}. This provides an argument that the optimal solution of GCC-SSP is attainable in the discretized GCC-SSP.

Without loss of generality, we assume $C^j_{\max} \le P^j$. 
The first direction follows from Eqn.~\raf{eq:disc} and the fact that we round up the cost function. Given a run  $s_0,s_1,...,s_h$ that satisfies $\sum_{t=0}^{h-1} \hat  C^j(s_t, \hat \pi(s_t)) \le  (1+\epsilon) P^j$, we have 
\begin{align*}
\sum_{t=0}^{h-1} C^j(s_t, \hat \pi(s_t) ) \le K^j \sum_{t=0}^{h-1} \hat C^j(s_t, \hat \pi(s_t) ) \le (1+\epsilon) P^j.
\end{align*}
Hence, $\Pr(\sum_{t=0}^{h-1}\hat  C^j(S_t, \hat \pi(S_t) ) \le (1+\epsilon) P^j)  \le \Pr(\sum_{t=0}^{h-1} C^j(S_t, \hat \pi(S_t) ) \le (1+\epsilon) P^j)$, which implies that
\begin{equation}
\begin{aligned}
     &\Pr\Big(\sum_{t=0}^{h-1} C^j(S_t, \hat \pi(S_t) ) > (1+\epsilon) P^j\Big) \\
     &\hspace{5mm}  \le \Pr\Big(\sum_{t=0}^{h-1}\hat  C^j(S_t, \hat \pi(S_t) ) > (1+\epsilon) P^j\Big) \le \Delta^j.
\end{aligned}
\end{equation}
Conversely, given a run $s_0,s_1,...,s_h$ that satisfies $\sum_{t=0}^{h-1}   C^j(s_t,  \pi'(s_t)) \le  P^j$, we have 
\begin{align}
& K \sum_{t=0}^{h-1} \hat C^j(s_t, \pi'(s_t)) \le    \sum_{t=0}^{h-1}  C^j(s_t, \pi'(s_t)) + hK \notag\\
&\hspace{30mm} \le P^j +  h \frac{\epsilon C^j_{\max }}{h}  \le (1+\epsilon) P^j,
\end{align}
where we use the property $a \lceil \frac{b}{a} \rceil \le b + a$ for $a,b\in \RR_+$. Therefore,
\begin{align}
&\Pr\Big(\sum_{t=0}^{h-1}   C^j(s_t,  \pi'(s_t)) \le  P^j \Big) \notag\\&\hspace{14mm} \le \Pr\Big(K \sum_{t=0}^{h-1} \hat C^j(s_t, \pi'(s_t)) \le  (1+\epsilon) P^j\Big)\\
&\iff \Pr\Big(K \sum_{t=0}^{h-1} \hat C^j(s_t, \pi'(s_t)) > (1+\epsilon) P^j\Big) \notag\\&\hspace{15mm} \le  \Pr\Big(\sum_{t=0}^{h-1}   C^j(s_t,  \pi'(s_t)) >  P^j\Big) \le \Delta^j.
\end{align}

\begin{corollary}
GCC-SSP under resource augmentation (Cons.~\raf{eq:aug}) can be reduced in polynomial time to CC-SSP. Therefore, any efficient algorithm that solves CC-SSP can be invoked  to solve GCC-SSP efficiently under resource augmentation.
\end{corollary}
The result follows directly from Lemma~\ref{lem:red} and Theorem~\ref{thm:dis}.

\section{Experiment}

\begin{figure}[ht]
    \centering
    \includegraphics[scale=0.5]{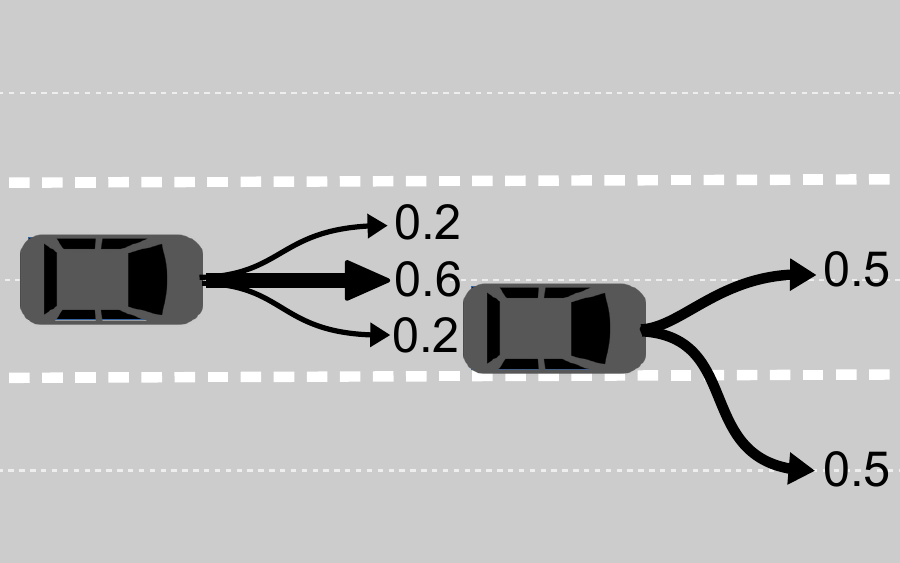}
    \caption{The transition function of the human-driven vehicles.}
    \label{fig:hv-t}
\end{figure}

To test the performance of the proposed CC-SSP model, we utilize two benchmark problems. The first problem is a two-dimensional grid problem where a robot can move in four directions. However, the movement is uncertain with an 80\% success probability captured in the transition function. A ratio of 5\% of the states are randomly selected as risky states, and 10\% of the states are randomly assigned a cost of $1$ (for all actions), and the remaining states assigned a cost of $2$. The grid size is $10000\times 10000$ (chosen to illustrate the CC-SSP's performance in a large state problem), and the initial state is set to $(5000, 5000)$.
The second problem is the highway problem (SSP version of the problem \cite{khonji2020risk}), where an Autonomous Vehicle (AV) navigates in a three-lane highway with multiple dynamic Human-driven Vehicles (HVs). The HVs move based on a transition probability depicted in Fig. \ref{fig:hv-t}. We consider that an HV deviates from the center of a lane before executing a change lane maneuver. A risky state is defined when the ego-vehicle collides with any of the agent vehicles.
The cost function for AVs actions (defined as maintain, speed up, slow down, left lane, right lane) is $(2, 1, 4, 3, 3)$, respectively. Both problems have a cost-minimizing objective. Moreover, the initial state consists of the AV and six HVs (see Fig.~\ref{fig:hw}).
The experiments were conducted  on an Intel i9 9900k processor using Gurobi 9 optimizer. 
Moreover, in Table \ref{tab:my-table}, the tree-based nodes represents the number of nodes in the And-Or tree history expansion similar to the approach used in \cite{sungkweon2021ccssp,huang2018hybrid,huang2019online}. 

\subsection{Results}
Table \ref{tab:my-table} demonstrates the CC-SSP solvers using the ILP formulation on the two benchmark problems. The tree-based nodes represents the number of nodes in the And-Or tree history expansion similar to the approach used in \cite{sungkweon2021ccssp,huang2018hybrid,huang2019online}. Both problems were solved for multiple numbers of horizons and risk thresholds ($\Delta$). The higher risk threshold results in a better objective, which is expected considering that the risk constraint is less restrictive. Moreover, while the objective increases proportionally with the horizon proportional, the objective value to horizon ratio is decreasing, indicating a better average policy per step is found considering the longer planning horizon. Besides, the CC-SSP only explores a fraction of the total number of nodes, allowing for a longer planning horizon in a short running time. While the ILP produces an optimal solution, the randomized rounding algorithm returns a close-to-optimal solution in a fraction of the time. 
Fig.~\ref{fig:aprox_confid_interval} plots the confidence interval of the randomized rounding algorithm's approximation ratio (computed as the ratio between the algorithm's objective value to that of an optimal solution) applied to the grid problem and repeated 100 times.
The Randomized Rounding algorithm performs well with a minimum approximation ratio of 0.94 of the optimal solution.

\begin{table*}[!htb]
    \centering
    \begin{minipage}{.6\textwidth}
        \footnotesize
        \centering
        \resizebox{1\textwidth}{!}{%
        \begin{tabular}{|l|l|l|l|l|l|l|l|l|}
        \hline
        & Horizon & $\Delta$ & \begin{tabular}[c]{@{}c@{}}ILP\\ Objective\end{tabular} & \begin{tabular}[c]{@{}c@{}}ILP Time\\  (sec)\end{tabular} & \begin{tabular}[c]{@{}l@{}}Rounding\\ Objective\end{tabular} & \begin{tabular}[c]{@{}l@{}}Rounding\\ Time (sec)\end{tabular} & \begin{tabular}[c]{@{}l@{}}Graph-based\\ \# nodes\end{tabular} & \begin{tabular}[c]{@{}l@{}}Tree-based\\ \# nodes\end{tabular} \\ \hline
        \multirow{8}{*}{\begin{tabular}[c]{@{}l@{}}Grid Problem\\ (10000x10000)\end{tabular}}    & \multirow{2}{*}{$h=10$} & 5\%    & 13.603   & 0.289    & 13.825     & 0.0637     & 506           & 5.36x$10^{9}$   \\ \cline{3-9} 
        &                         & 10\%     & 12.958       & 0.2448         & 13.824            & 0.064              & 506              & 5.36x$10^{9}$   \\ \cline{2-9} 
           & \multirow{2}{*}{$h=25$} & 5\%      & 30.975       & 8.983          & 31.743            & 4.492              & 6,201            & 8.67x$10^{25}$  \\ \cline{3-9} 
           &                         & 10\%     & 29.414       & 11.315         & 31.743            & 4.257              & 6,201            & 8.67x$10^{25}$  \\ \cline{2-9} 
           & \multirow{2}{*}{$h=30$} & 5\%      & 36.767       & 50.790         & 37.716            & 12.758             & 10,416           & 2.16x$10^{31}$  \\ \cline{3-9} 
           &                         & 10\%     & 34.902       & 43.099         & 37.716            & 13.299             & 10,416           & 2.16x$10^{31}$  \\ \cline{2-9} 
           & \multirow{2}{*}{$h=35$} & 5\%      & 42.555       & 126.046        & 43.687            & 29.072             & 16,206           & 5.37x$10^{36}$  \\ \cline{3-9} 
           &                         & 10\%     & 40.386       & 101.244        & 43.687            & 41.808             & 16,206           & 5.37x$10^{36}$  \\ \hline
        \multirow{8}{*}{\begin{tabular}[c]{@{}l@{}}Highway Problem\\ (Three lanes)\end{tabular}} & \multirow{2}{*}{$h=4$}  & 5\%      & 6.258        & 0.165          & 6.448             & 0.098              & 2,438            & 124,251       \\ \cline{3-9} 
           &                         & 10\%     & 5.854        & 0.148          & 5.854             & 0.090              & 2,438            & 124,251         \\ \cline{2-9} 
           & \multirow{2}{*}{$h=5$}  & 5\%      & 7.343        & 1.274          & 7.642             & 0.356              & 7,666            & 1.99x$10^{6}$   \\ \cline{3-9} 
           &                         & 10\%     & 6.833        & 0.655          & 6.833             & 0.337              & 7,666            & 1.99x$10^{6}$   \\ \cline{2-9} 
           & \multirow{2}{*}{$h=6$}  & 5\%      & 8.343        & 4.249          & 8.712             & 1.235              & 20,576           & 3.18x$10^{7}$   \\ \cline{3-9} 
           &                         & 10\%     & 7.761        & 2.721          & 7.762             & 1.247              & 20,576           & 3.18x$10^{7}$   \\ \cline{2-9} 
           & \multirow{2}{*}{$h=7$}  & 5\%      & 9.317        & 20.598         & 9.759             & 3.929              & 49,526           & 5.09x$10^{9}$   \\ \cline{3-9} 
           &                         & 10\%     & 8.672        & 10.552         & 8.672             & 3.444              & 49,526           & 5.09x$10^{9}$   \\ \hline
        \end{tabular}
        }
        \vspace{-5pt}
        \caption{The CC-SSP solvers comparison on two benchmark problems (with minimization objective functions).}
        \label{tab:my-table}
    \end{minipage}%
    ~
    \begin{minipage}{0.4\textwidth}
        
            \centering
               \includegraphics[scale=0.45]{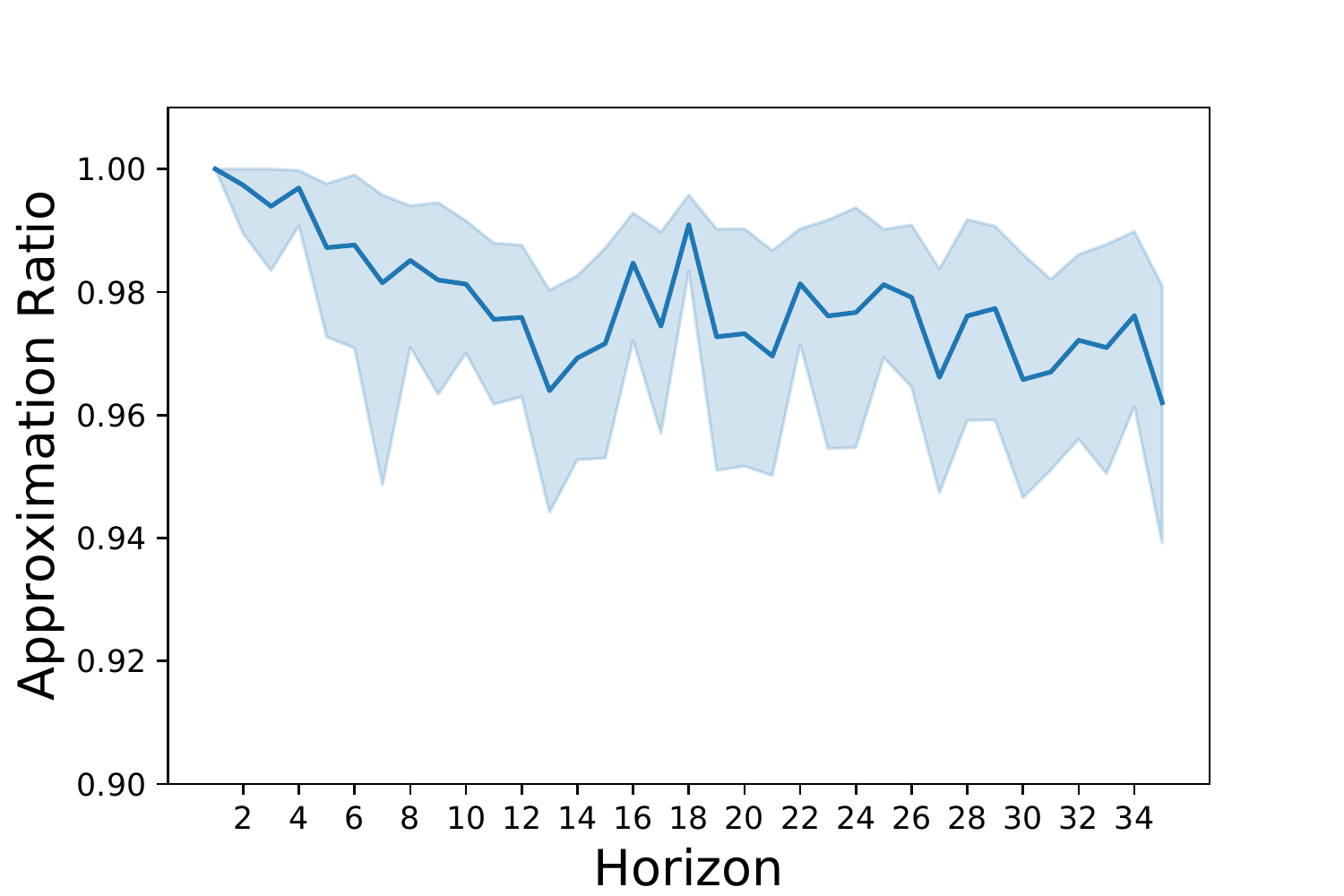}\vspace{-5pt}
            \captionof{figure}{The approximation ratio of the randomized rounding algorithm, applied to the grid problem, at $99\%$ confidence interval.}
            \label{fig:aprox_confid_interval}
    \end{minipage}
\end{table*}

\section{Conclusion}
We presented a novel formulation for CC-SSP under local and global risk constraints. To optimize the running time, we provided an iterative rounding algorithm to obtain feasible policies efficiently. Experiments show that our approaches offer a significant improvement over existing techniques that rely on history tree expansion. In future work, we intend to find an approximation algorithm for the ILP that runs in polynomial time and provides a theoretical guarantee on the sub-optimality ratio.


\bibliographystyle{abbrv}
\bibliography{reference}

\section{Appendix}

\subsection{Proof of Lemma~\ref{lem:rec}}
\begin{proof}
The execution risk can be written as,
\footnotesize{}
\begin{align}
&\textsc{Er}^j(s_k) = 1 -  \Pr\Big(\bigwedge_{t=k}^{h} \neg R^j(S_t) \mid S_k = s_k\Big)  \notag \\
&= 1 -  \bigg(\Pr\Big(\bigwedge_{t=k}^{h} \neg R^j(S_t) \mid S_k = s_k, \neg R^j(S_k) \Big) \notag \\
& \hspace{5mm}\cdot \Pr(\neg R^j(S_k) \mid S_k = s_k) \notag  + \Pr(R^j(S_k) \mid S_k = s_k) \\
& \hspace{25mm}\cdot \Pr\Big(\bigwedge_{t=k}^{h} \neg R^j(S_t) \mid S_k = s_k,  R^j(S_k) \Big)  \notag \bigg)\\
&= 1 - \Pr\Big(\bigwedge_{t=k+1}^{h} \neg R^j(S_t) \mid S_k = s_k, \neg R^j(S_k) \Big) (1-r^j(s_k)). \label{eq:er1}
\end{align}
The probability term can be expanded, conditioned over subsequent states at time $k+1$,
\begin{align}
&\sum_{s_{k+1}\in \cS} \Pr\Big(\bigwedge_{\mathclap{t=k+1}}^{h} \neg R^j(S_t) \mid S_k = s_k, \neg R^j(S_k), S_{k+1} = s_{k+1} \Big) \notag \\& \hspace{30mm} \cdot \Pr(S_{k+1} = s_{k+1} \mid S_k = s_k, \neg R^j(S_k)) \notag \\
&= \sum_{s_{k+1}} \Pr\Big(\bigwedge_{t=k+1}^{h} \neg R^j(S_t) \mid S_{k+1} = s_{k+1}\Big) \notag \\& \hspace{35mm} \cdot \Pr(S_{k+1} = s_{k+1} \mid S_k = s_k) \label{eq:ind'}  \\
& = \sum_{s_{k+1}} (1 - \textsc{Er}^j(s_{k+1}) )  T(s_k, \pi(s_k), s_{k+1}), \label{eq:er2}
\end{align}
\normalsize{}
where Eqn.~\raf{eq:ind'} follows from the independence between $\big(\bigwedge_{t=k+1}^{h} \neg R^j(S_t) \mid S_{k+1}= s_{k+1}\big)$ and $(S_k = s_k\wedge \neg R^j(S_k))$, by the Markov property, and between $(S_{k+1} = s_{k+1} \mid S_k = s_k)$ and $\neg R^j(S_k)$.
Combining Eqns.~\raf{eq:er1},\raf{eq:er2} we obtain $\textsc{Er}^j(s_k)$ as

\footnotesize{}
\begin{align}
&\textsc{Er}^j(s_k) =  1 - (1-r^j(s_k)) \sum_{s_{k+1}} \big( (1-\textsc{Er}^j(s_{k+1}) )   T(s_k, \pi(s_k), s_{k+1}) \big) \notag \\
&= 1 +(1-r^j(s_k))  \Big(\sum_{s_{k+1}} \big(\textsc{Er}^j(s_{k+1}) T(s_k, \pi(s_k), s_{k+1})\big)  \notag \\& \hspace{45mm} - \sum_{s_{k+1}}   T(s_k, \pi(s_k), s_{k+1}) \Big) \notag \\
&= 1 +(1-r^j(s_k))  \sum_{s_{k+1}} \big( \textsc{Er}^j(s_{k+1})T(s_k, \pi(s_k), s_{k+1}) \big)  - (1-r^j(s_k))\notag \\
&=r^j(s_k) + (1-r^j(s_k))  \sum_{s_{k+1}} \big( \textsc{Er}^j(s_{k+1}) T(s_k, \pi(s_k), s_{k+1}) \big) \notag \\
&= r^j(s_k) + \sum_{s_{k+1}} \big(  \textsc{Er}^j(s_{k+1}) \widetilde T^j(s_k, \pi(s_k), s_{k+1})\big) \label{eq:er3}
\end{align}
\normalsize{}
Note that the execution risk of $\textsc{Er}^j(s_{h}) = r^j(s_{h})$. For a stochastic policy $\pi$, Eqn.~\raf{eq:er3} can be rewritten by enumerating through all actions, which completes the proof. 

\end{proof}

\subsection{Proof of Theorem~\ref{lem1} }
\begin{proof}
We can rewrite the execution risk in Eqn.~\raf{eq:exec-risk} as $\textsc{Er}'^j(s_k) =$
\begin{equation}
 \left\{\begin{array}{l}
 r^j(s_0) + \sum_{s_{k+1}\in \cS}\sum_{a \in \cA} ( \textsc{Er}'^j(s_{k+1}) + r^j(s_{k+1}) ) \notag \\
 \hspace{10mm} \cdot \pi(s_k, a) \widetilde T^j(s_k, a, s_{k+1}) \hspace{6mm} \text{if } k = 0, \\
\sum_{s_{k+1}\in \cS}\sum_{a \in \cA} ( \textsc{Er}'^j(s_{k+1}) + r^j(s_{k+1}) ) \notag\\
 \hspace{10mm} \cdot \pi(s_k, a) \widetilde T^j(s_k, a, s_{k+1})\hspace{6mm} \text{if } k = 1,...,h-1,\\
0 \hspace{49mm} \text{if } k=h.
\end{array}\right. \label{eq:er0}
\end{equation}
where $\textsc{Er}^j(s_k) = \textsc{Er}'^j(s_k)  +  r^j(s_k)$, and $\textsc{Er}^j(s_0) = \textsc{Er}'^j(s_0)$.
Based on the  flow equations \raf{conf1},\raf{conf2}, define a policy 
\begin{equation}
\pi(s_k,a) := \left\{ \begin{array}{l l}
 x^j_{s,0,a} \hspace{30pt}\vspace{10pt},& \text{if } s_k = s_0 \\
 \frac{x_{s,k,a}^j}{\sum_{a' \in \cA}  x_{s,k,a'}^j}, & \text{otherwise.}
\end{array}\right.
\end{equation}
Note that the policy is a valid probability distribution. Thus, we rewrite Eq.~\raf{conf1},\raf{conf2} by
\begin{align}
    & x^j_{s,k,a} = \pi(s_k,a) \sum_{s_{k-1}\in\cS}\sum_{a'\in \cA} x^j_{s,k-1,a'} \widetilde T^j(s_{k-1},a',s_k) \notag \\
    & x^j_{s,0,a} = \pi(s_0,a). \label{eq:pix}
\end{align}
Next, we proof by induction the following statement
\begin{align}
&\textsc{Er}^j(s_0) = r^j(s_0) \notag \\ & \hspace{5mm} + \sum_{k = 1}^{h'}\sum_{s_{k}\in \cS}\sum_{a \in \cA} \sum_{s_{k-1}\in \cS}  r^j(s_{k}) x^j_{s,k-1,a} \widetilde T^j(s_{k-1}, a, s_{k}) \notag \\
&\hspace{5mm} +  \sum_{s_{h'}\in \cS}\sum_{a \in \cA} \sum_{s_{h'-1}\in \cS}\textsc{Er}'^j(s_{h'})  x^j_{s,h'-1,a} \widetilde T^j(s_{h'-1}, a, s_{h'}). \label{eq:ind}
\end{align}
Note that when $h' = h$, by Eqn.~\raf{eq:er0}, the last term of the above equation is zero, which is equivalent to the lemma's claim.
We consider the initial case with $h'=1$. From Eqn.~\raf{eq:er0} and $\textsc{ER}^j(s_0) = \textsc{ER}^{'j}(s_0)$, we obtain
\begin{align*}
    &\textsc{Er}^j(s_0) = r^j(s_0) \notag \\ & \hspace{5mm} + \sum_{s_{1}\in \cS}\sum_{a \in \cA} ( r^j(s_{1}) + \textsc{Er}'^j(s_{1})  ) \pi(s_0, a) \widetilde T^j(s_0, a, s_{1}) \\ 
     &= r^j(s_0) + \sum_{s_{1}\in \cS}\sum_{a \in \cA} r^j(s_{1}) x_{s,0,a} \widetilde T^j(s_0, a, s_{1}) \notag \\ & \hspace{14mm} + \sum_{s_{1}\in \cS}\sum_{a \in \cA} \textsc{Er}'^j(s_{1}) x_{s,0,a} \widetilde T^j(s_0, a, s_{1}),
\end{align*}
where $s_{h'-1} = s_0$ is a known state.
For the inductive step, we assume  Eqn.~\raf{eq:ind} holds up to $h'=i$, we proof the statement for $h'=i+1$. Expanding $\textsc{Er}^{'j}(s_i)$ using Eqn.~\raf{eq:er0} obtains
\begin{align}
 &\textsc{Er}^j(s_0) = r^j(s_0) \notag \\ & \hspace{5mm} + \sum_{k=1}^i \sum_{s_{k}\in \cS}\sum_{a \in \cA} \sum_{s_{k-1}\in \cS} \Big( r^j(s_{k}) x_{s,k-1,a} \widetilde T^j(s_{k-1}, a, s_{k}) \Big) \notag \\
 &\hspace{5mm}  + \sum_{s_{i}\in \cS}\sum_{a \in \cA}\sum_{s_{i-1}\in \cS} \big( \textsc{Er}'^j(s_{i}) x_{s,i-1,a} \widetilde T^j(s_{i-1},a,s_i) \big) \notag
 \end{align}
    \begin{align}
     &= r^j(s_0) \notag \\ & \hspace{5mm}+ \sum_{k=1}^i \sum_{s_{k}\in \cS}\sum_{a \in \cA}\sum_{s_{k-1}\in \cS} \Big( r^j(s_{k}) x_{s,k-1,a} \widetilde T^j(s_{k-1}, a, s_{k}) \Big) \notag\\
    & \hspace{5mm} + \sum_{s_{i+1}\in \cS}\sum_{a' \in \cA}\sum_{s_{i}\in \cS} \Big( \big(  r^j(s_{i+1}) + \textsc{Er}'^j(s_{i+1})  \big)\pi(s_i,a') \notag \\ & \hspace{5mm} \cdot  \widetilde T^j(s_i,a',s_{i+1})  \sum_{s_{i-1}\in \cS}\sum_{a \in \cA} \big(x_{s,i-1,a}   \widetilde T^j(s_{i-1}, a, s_{i} \big)  \Big) \notag
    \end{align}
    \begin{align}
     &= r^j(s_0) \notag\\& +  \sum_{k=1}^i \sum_{s_{k}\in \cS}\sum_{a \in \cA}\sum_{s_{k-1}\in \cS} \Big( r^j(s_{k}) x_{s,k-1,a} \widetilde T^j(s_{k-1}, a, s_{k}) \Big) \notag \\
     &  + \sum_{s_{i+1}\in \cS}\sum_{a \in \cA}\sum_{s_{i}\in \cS} \Big( \big(  r^j(s_{i+1}) + \textsc{Er}'^j(s_{i+1})  \big) \notag\\&\hspace{40mm} \cdot x_{s,i,a} \widetilde T^j(s_i,a,s_{i+1}) \Big) \label{eq:simpx}  \end{align}
where Eqn.~\raf{eq:simpx}  follows by substituting $\pi(s_i,a')$, using Eqn.~\raf{eq:pix}, by
$$ \pi(s_i,a')  = \frac{ x^j_{s,i,a'}} {\sum_{s_{i-1}\in\cS}\sum_{a\in \cA} x^j_{s,i-1,a} \widetilde T^j(s_{i-1},a,s_i)}.$$
Rewriting Eqn.~\raf{eq:simpx} obtains
\begin{align}
&\textsc{Er}^j(s_0)= r^j(s_0) \notag\\&\hspace{2mm} + \sum_{k=1}^{i+1}\sum_{s_{k}\in \cS}\sum_{a \in \cA}\sum_{s_{k-1}\in \cS} r^j(s_{k}) x_{s,k-1,a} \widetilde T^j(s_{k-1}, a, s_{k})\notag\\
& \hspace{2mm} + \sum_{s_{i+1}\in \cS}\sum_{a \in \cA}\sum_{s_{i}\in \cS} \textsc{Er}'^j(s_{i+1})   x_{s,i,a} \widetilde T^j(s_i,a,s_{i+1}),
\end{align}
\end{proof}

\color{black}

\end{document}